\documentclass[runningheads]{llncs}
\usepackage{graphicx}

\usepackage{algpseudocode}
\usepackage{algorithm}

\makeatletter
\renewcommand{\ALG@beginalgorithmic}{\small}
\makeatother

\usepackage{xcolor}

\usepackage{amssymb}

\begin{document}

\title{Real-Time Hand Gesture Identification in Thermal Images}


\author{James M. Ballow\inst{1}\orcidID{0000-0003-0548-9901} \and
Soumyabrata Dey\inst{1}\orcidID{0000-0002-4589-5165}}

\authorrunning{J. Ballow and S. Dey}

\institute{Clarkson University, 8 Clarkson Avenue, Potsdam, NY 13699, USA \email{ballowjm@clarkson.edu; sdey@clarkson.edu}}

\maketitle              

\begin{abstract}
    Hand gesture-based human-computer interaction is an important problem that is well explored using color camera data. In this work we proposed a hand gesture detection system using thermal images. Our system is capable of handling multiple hand regions in a frame and process it fast for real-time applications. Our system performs a series of steps including background subtraction-based hand mask generation, k-means based hand region identification, hand segmentation to remove the forearm region, and a Convolutional Neural Network (CNN) based gesture classification. Our work introduces two novel algorithms, \textit{bubble growth} and \textit{bubble search}, for faster hand segmentation. We collected a new thermal image data set with 10 gestures and reported an end-to-end hand gesture recognition accuracy of $\sim97\%$.          

\end{abstract}

    
    \begin{figure} [t]
        \includegraphics[width=\textwidth]{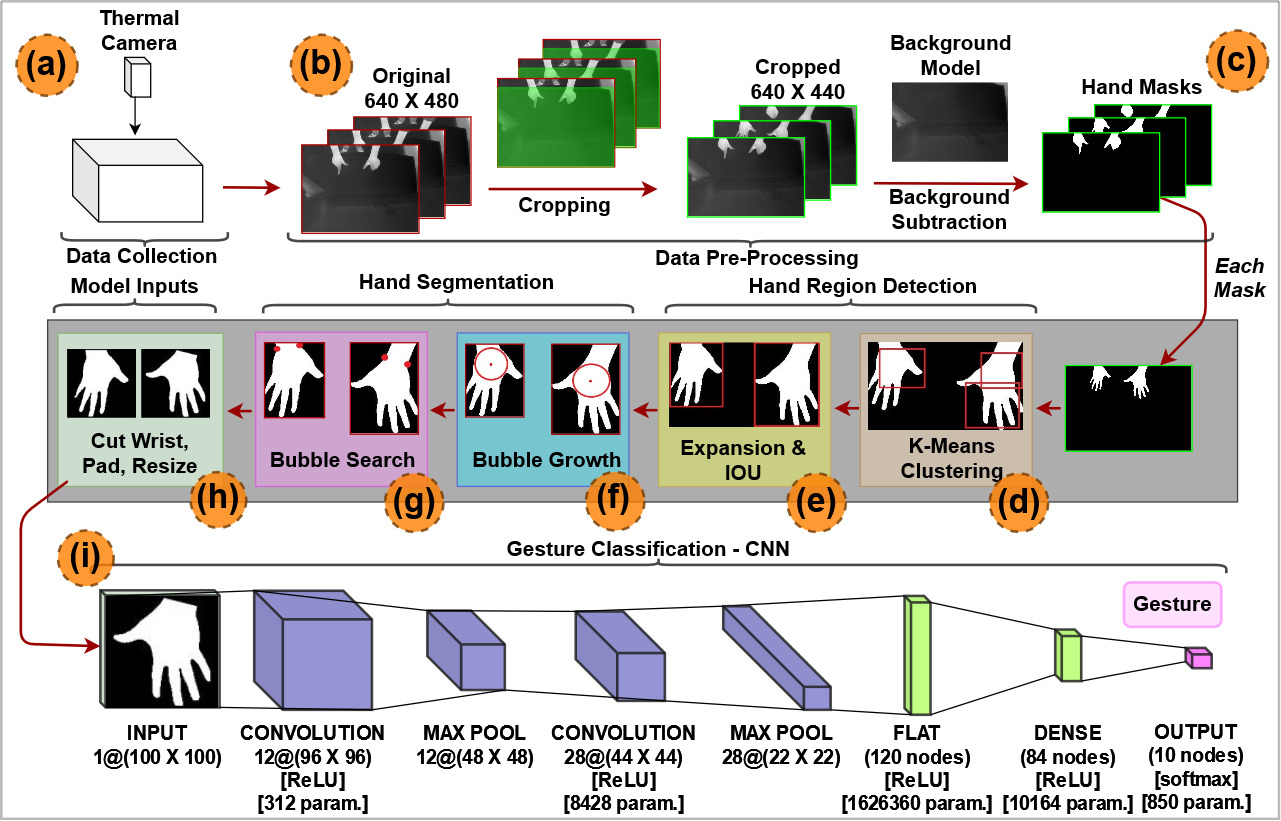}
        
        \caption{End-to-end flowchart for (a) data capture, (b) frame cropping and background subtraction, (c) hand mask generation, (d-e)  hand region detection, (f-g) hand segmentation using key-points, (e) final model input formatting, (i) and gesture classification.}
        
        \label{figure:Overall Flowchart}
        \vspace{-0.2in}
    \end{figure}

\vspace{-0.2in}
\section{Introduction}

    Communication between human and computer via hand gestures has been studied extensively and continues to be a fascination in the computer vision community. This is not a surprise given the proliferation in use of artificial intelligence over the past decade to improve the lives of many through the development of \textit{smart} systems. For humans to convey intention to computers, studies began with the use of external devices \cite{ColorBased_TabletopInterfaces,LayeredAchitecture,PoseRecovery_UsingCNN,handTracking_LowCostHardware} to enhance focus on particular regions of hands so as to limit the number of features required to interpret hand movement sequences into pre-determined messages. Over time, more advanced equipment was introduced to eliminate the need for external devices to be attached to a user's hands to isolate hand movements; equipment like the use of a depth camera \cite{FastRobust_DetectionOptimization,DeepDynamic_SegmentationRecognition,COP_WristSegementation_2013}, high resolution RGB cameras \cite{RealTimeHandGesture_2014,RealTime_BagOfFeatures,Purple_Web_Hands}, and (less-frequently) thermal cameras \cite{AHD}. 

    Even though hand gesture detection is a well-explored problem using some modalities of data such as RGB camera images, there have been a very limited number of works on thermal data \cite{In-Air,FastTrackingHandsFingertips,AHD}. These studies have typically required multiple sensors, a fixed location of the hand in frame to estimate wrist points, or have a user wear clothing to separate the hand from the forearm.
    
    Thermal modality can complement RGB data modality because it is not affected by different lighting conditions and skin color variance. Moreover, thermal data based analyses can be extended to swipe detection techniques by temperature tracing on natural surfaces \cite{materialDetection}. Therefore, future research can benefit by using multi-modal (thermal and RGB) data for a robust hand gesture detection system. To this end, through research on hand gesture detection techniques using thermal data is essential.
    
    Efficient hand segmentation is vital to the success of thermal camera based hand detection. This is because thermal images lack many distinguishable features such as color and textures, and including regions from other heated objects such as a forearm can reduce the classification accuracy. Our algorithmic pipeline uses background subtraction in the data pre-processing stage for generating a hand mask, k-means clustering and overlapping cluster grouping for hand region isolation of each potential hand region, center of palm and wrist point detection for hand segmentation (removing forearm), and a CNN-based model gesture classification (Figure \ref{figure:Overall Flowchart}). In this process we introduced two novel algorithms, \textit{bubble growth} and \textit{bubble search}, for hand segmentation. The main contributions of this paper are as follows: 1) Collection of a new thermal hand gesture data set from 23 users performing the same 10 gestures. 2) \textit{Bubble Growth} method, which uses a distance transform and hand-forearm contour to expand a circle (bubble) to the maximum extent possible inside the hand. 3) \textit{Bubble Search} method, which was inspired by the use of an expansion of the maximum inscribed circle and a threshold distance of two consecutive contour points as detailed in \cite{RealTimeHandGesture_2014}, but includes additional constraints, a reference point, and an evolving (in lieu of fixed) bubble expansion. 4) Developing an end-to-end real-time hand gesture detection system that can process multiple hands at a frame-rate of 8 to 10 frames per second (fps) with high accuracy ($\sim97\%$.). Our \textit{bubble growth} and \textit{bubble search} methods are superior to other methods because they neither use nor require knowledge of any projection \cite{HandLandmarks,Purple_Web_Hands}, angle of hand rotation \cite{Purple_Web_Hands}, finger location or fixed values for bubble radii \cite{RealTimeHandGesture_2014,hand_pose_estimation}, degree of palm roundness \cite{COP_WristSegementation_2013}. The combined average speed of our methods (\textit{bubble growth} = 0.012 sec/hand, \textit{bubble search} = 0.007 sec/hand; total = 0.019 sec/hand).

\section{Methods}
    This paper proposes a process (Figure \ref{figure:Overall Flowchart}) that can perform real-time hand detection and gesture classification from a thermal camera video feed. The process is divided into 5 major parts: (1) data collection; (2) data pre-processing; (3) hand region detection; (4) hand segmentation, and (5) gesture classification.

    \begin{figure} [t]
    \begin{center}
        \includegraphics[width=1.0\linewidth]{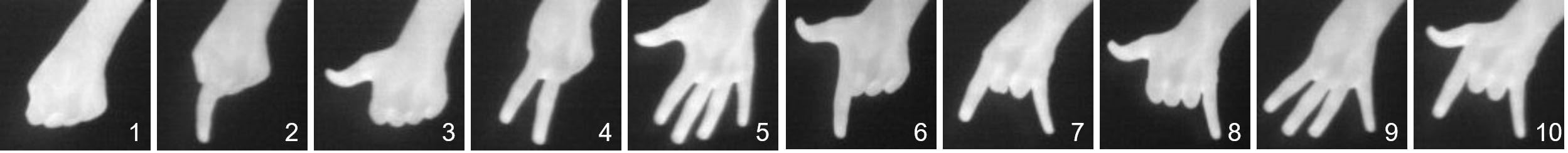}
        \caption{Gestures to test the proposed algorithms and train our CNN model.} 
        \label{figure: All Ten Gestures}
        \vspace{-0.1in}
    \end{center}    
    \end{figure}

    \begin{table} [t]
        \centering
        \caption{Data set composition by gesture number ($G_x$) and left/right hands.}
        \label{Table: Data Collection Information}
        \begin{tabular}{| l | c | c | c | c | c | c | c | c | c | c | c | c | c |}
            \hline
            {}      & Users & $G_1$ & $G_2$ & $G_3$ & $G_4$ & $G_5$ & $G_6$ & $G_7$ & $G_8$ & $G_9$ & $G_{10}$ & Left & Right \\
            \hline
            Training Data        & 20 & 1536 & 1576 & 1438 & 1544 & 1183 & 1459 & 1358 & 1391 & 1524 & 1285 & 6657 & 7637  \\
            \hline
            Test Data            & 3  & 103	& 57 & 34 & 92 & 115 & 30 & 17 & 92 & 138 & 214 & 517 & 355 \\
            \hline
        \end{tabular}
        \vspace{-0.1in}
    \end{table}

\vspace{-0.1in}
\subsection{Data Collection}
    All thermal hand gesture video data is collected with a Sierra Olympic Viento-G thermal camera with a 9mm lens. The video frames are recorded in indoor conditions (temperature between $65^{\circ}$ to $70^{\circ}$ F) at 30 fps and stored as 16-bit TIFF images with a $640 \times 480$ pixel resolution. The camera is fixed to a wooden stand and oriented downwards towards a tabletop. 
    
    Data was collected from 23 users demonstrating 10 pre-defined gestures with left and right hands to develop training and test data sets. Separate sets of users contributed to the training and test data sets to demonstrate that our method is user-agnostic. Figure \ref{figure: All Ten Gestures} illustrates all 10 gestures, and Table \ref{Table: Data Collection Information} lists the data we have collected and divided into a training data set, testing data set. 
    
    For this paper, we also used an external data called \textit{Finger Digits 0-5} \cite{FingerDigits} set to assess how our process generalizes. From this set we tested gesture 1, 2, 4, 5, and 9 (2000 images per gesture). This data set was not used in training or in any other experimentation in this project.
   
\subsection{Data Pre-processing}
    Without loss of generality to image size, we reduced the size of our images from $640 \times 480$ to $640 \times 440$ to reduce camera scope to the boundaries of the table top to make background subtraction simpler. Images were also converted to 8-bit JPG images for ease of viewing and using certain python packages (e.g., \textit{OpenCV}). 
    
    A MOG2 background subtraction model from \textit{OpenCV} was used to generate hand masks from each thermal image. We initialized the model with frames that, at the beginning of data capture, contain no hands or heated objects and have a table at constant room temperature. The model is updated over the video sequence when there are no pixels found in the hand mask, essentially allowing only pixels associated with slight change in room temperature to be updated. Each mask is binarized (black and white pixels) using Otsu's method to select an appropriate threshold value.

    \begin{figure} [t]
        \includegraphics[width=\textwidth]{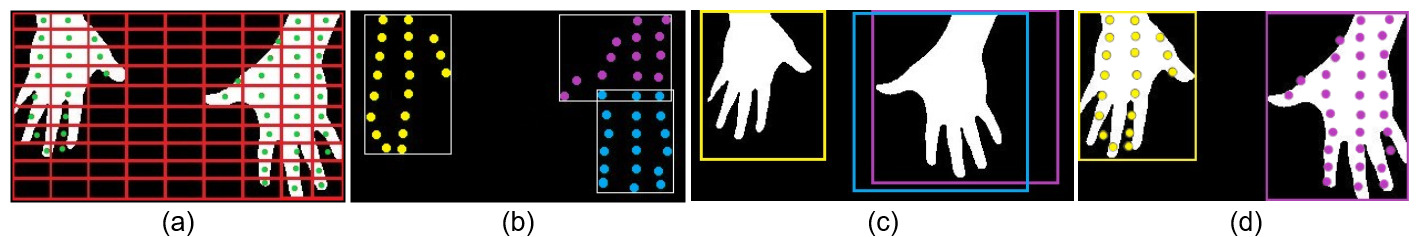}
        \caption{Hand region detection: (a-b) centers of white pixel in all grids are clustered using k-means. (c-d) optimal number of hand regions are detected.}
        \label{figure:Kmeans Clustering}
        \vspace{-0.15in}
    \end{figure}

\vspace{-0.1in}    
\subsection{Hand Region Detection}
    Hand regions are identified in hand masks using tightly bounding boxes around each hand object (which may or may not contain any length of forearm) using a two-step process detailed below.

    First, a k-means clustering algorithm (Figure \ref{figure:Kmeans Clustering}) is used to identify contiguous objects in the hand mask using a number of centroids estimated with a silhouette analysis optimal cluster-finding technique. To achieve real-time processing speed, we expedite this step by reducing the number of points to consider when clustering. To do this, we encapsulate all white pixels in the hand mask with a grid that is subdivided into equally-sized partitions. For each coupon containing at least one white pixel, the center of mass is calculated to reduce a coupon's points down to a single point. The silhouette analysis \cite{SilhouetteAnalysis} identifies the optimal cluster number by performing k-means clustering for different $k$ values (k in the range of 2 to 3) and selecting the optimal $k$ for which the highest silhouette score is calculated. 
    
    Second, a bounding box is placed initially around each cluster but is expanded in all dimensions equally until the box entirely contains a set of contiguous white pixels in the hand mask. If the number of centroids selected is larger than the number of hands in the hand mask (e.g., poor hand mask generation), then one or more regions will be bounded by multiple boxes. Removal of these duplicate boxes is performed using intersection-over-union (IOU) and a threshold of 0.7---the boxes that remain after IOU are the hand regions.

    \begin{algorithm} [t]
        \caption{Center of Palm Detection}
        \label{algorithm:BubbleGrowth}
        \begin{flushleft}
            \textbf{Output:} \textit{$C_{COP}$, $R_{COP}$} 
            \begin{algorithmic}[1]
            \Function{BubbleGrowth}{$H_{part}$, $C_{est}$, $R_{est}$}
                \State $C_{cand}, R_{cand} \gets C_{est}, R_{est}$
                \State $C_{COP}, R_{COP} \gets C_{est}, R_{est}$
                \State $C_{visited} \gets \{\}$
                \Repeat 
                    \For {$h \gets H_{part}$}
                        \State $C_{cand} \gets \textproc{ShortAdvancement}(C_{COP},h,F_{pace})$
                        \If {$C_{cand}$ \textbf{is in} $C_{visited}$}
                            \State \textbf{continue}
                        \EndIf    
                        \State $R_{cand} \gets \textproc{ShortestDistance}(C_{cand},H_{part})$
                        \If {$R_{cand} > R_{COP}$} 
                            \State $C_{COP}, R_{COP} \gets C_{cand}, R_{cand}$
                            \State \textproc{SignalBubbleMoved()}
                            \State \textbf{break}
                        \EndIf
                        \State $C_{visited} \gets C_{cand}$
                    \EndFor
                \Until no \textproc{SignalBubbleMoved}
                \State \textbf{return} $C_{COP}, R_{COP}$
            \EndFunction
            \end{algorithmic}
        \end{flushleft}
        \vspace{-0.1in}
    \end{algorithm}

    \begin{algorithm} [h]
        \caption{Wrist Point Detection}
        \begin{flushleft}
        \textbf{Output:} $W_1, W_2$ \\
        \textbf{Requires:} $R_{exp}, D_{min}, D_{max}, i_{max}$
        \begin{algorithmic}[1]
        \Function{BubbleSearch}{$H_{all}$, $C_{COP}$, $R_{COP}$, $C_{Ref}$, $C_{Ref,edges}$}
            \State $D_{Ref} \gets \textproc{Distance}(C_{COP},C_{Ref})$ 
            \State $D_{max},D_{min},R_{exp} \gets \textproc{ApplyScalars}(R_{COP})$ 
            \State $L_{meetsCriteria} \gets \{\}$ 
            \textproc{}
            \If{$R_{COP} > D_{Ref}$}
                \State $W_1, W_2 \gets C_{Ref,edges}$
            \Else
                \Repeat
                    \State $H_{inside} \gets \textproc{PointsInsideCircle}(H_{all},C_{COP},R_{exp})$ 
                    \For{$i \gets \textproc{indicies}(H_{inside})$}
                        \State $P_1, P_2 \gets \textproc{GetElements}(H_{inside},i,i+1)$
                        \State $M_{1,2} \gets \textproc{Midpoint}(P_1,P_2)$
                        \State $D_1,D_2 \gets (\textproc{Distance}(P_1,P_2), \textproc{Distance}(M_{1,2},C_{Ref}))$
                        \If{$D_{min} < D_1 < D_{max}$ \textbf{and} $D_2 < D_{Ref}$} 
                            \State $L_{meetsCriteria} \gets ([D_1,D_2,P_1,P_2])$
                        \EndIf 
                    \EndFor
                    
                    \If {$\vert L_{meetsCriteria} \vert > 1$} 
                        \State $W_1, W_2 \gets \textproc{GetPointsWithSmallestD}_2(L_{meetsCriteria})$
                        \State \textbf{break}
                    \ElsIf {$i_{max}$ reached}
                        \State $W_1, W_2 \gets C_{Ref,edges}$
                    \Else
                        \State $R_{exp} \gets F_2 \times R_{exp}$ 
                    \EndIf
                \Until $i_{max}$ reached
            \EndIf 
            \State \textbf{return} $W_1, W_2$
        \EndFunction
        \end{algorithmic}
        \end{flushleft}
        \vspace{-0.15in}
        \label{algorithm:BubbleSearch}
    \end{algorithm}

\vspace{-0.15in}
\subsection{Hand Segmentation}

\begin{figure} [t]
        \centering
        \includegraphics[width=0.75\linewidth]{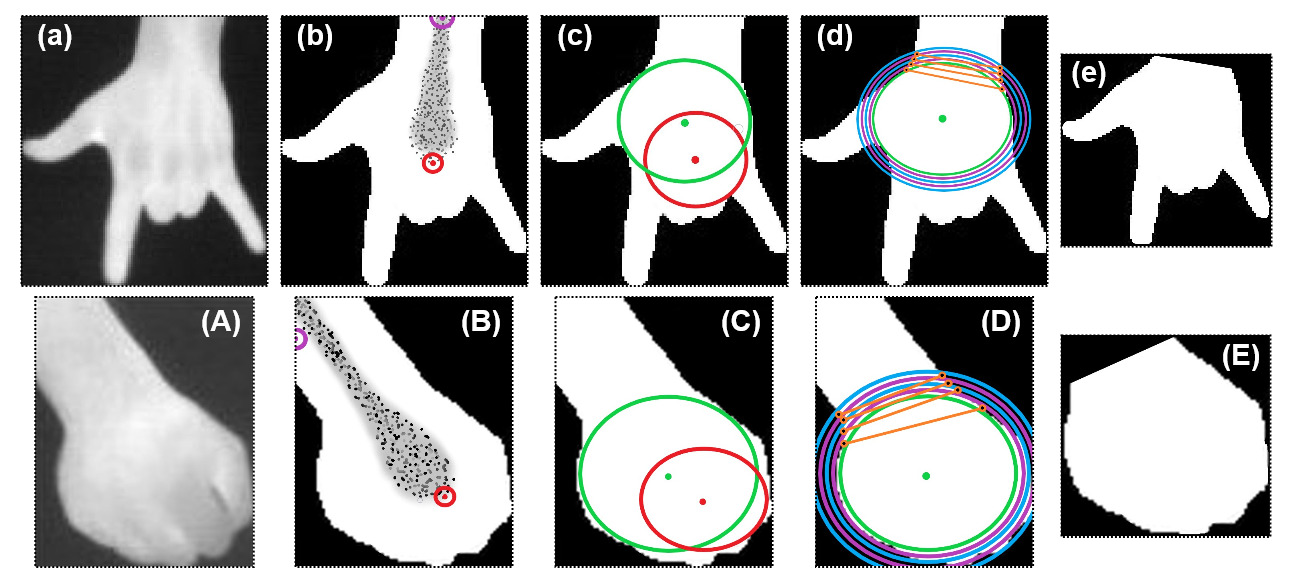}
        \caption{Hand segmentation with $C_{Ref}$ on a single edge of frame (a-e) and $C_{Ref}$ split between two edges (A-E). (a,A) original image; (b,B) DT and $C_{Ref}$ (purple) to find $C_{est}$ (red); (c,C) $C_{est}$ (red) and $C_{COP}$ (green); (d,D) $R_{exp}$ expanding to find appropriate $W_1, W_2$ (orange); (e,E) arm removed.} 
        \label{figure:Hand Segmentation}
        \vspace{-0.2in}
    \end{figure}

    A hand region may include variable forearm length which can hinder high-accuracy gesture classification due to lack of forearm agnosticism. This can be avoided by including gesture samples with variable lengths of forearms in the training set, but this would be a costly endeavor. Instead, as illustrated in Figure \ref{figure:Hand Segmentation} we algorithmically sever the hand region from the forearm at the wrist and thus removed all forearm-related data variation. The process uses two novel algorithms: Bubble Growth (to find COP) and Bubble Search (to find WP).  

    \textbf{Reference Point Determination}: Bubble Growth and Bubble Search require a reference point ($C_{ref}$) located on the edge through which the hand enters the frame, or \textit{hand penetration edge}. To get this point, we estimate the largest contiguous array of white pixels on each of the four edges of the hand region to be the hand penetration edge. The extreme ends of the array of white pixels on the hand penetration edges are set as the reference point edges ($C_{ref,edges}$) and the midpoint of these points yields $C_{ref}$.
    
    \textbf{Bubble growth:} Given a subset of contours ($H_{part}$) from the entire set of contours for the hand region ($H_{all}$) and an initial estimate for COP ($C_{est}$), Algorithm \ref{algorithm:BubbleGrowth} moves around the space in the palm to find the COP ($C_{COP}$). We form $H_{part}$ by taking a sparse subset of $H_{all}$ in a fashion that preserves the overall shape of the hand region; doing this optimized the cost of Algorithm \ref{algorithm:BubbleGrowth}. We compute $C_{est}$ by first calculating the distance transform (DT) \cite{DistanceTransform} of the entire hand region, obtaining the maximum DT value ($\eta_{max}$), and selecting all points in the hand region which have a DT value ($\eta$) such that $\eta \geq 0.80*\eta_{max}$. We select the point in this set furthest from $C_{ref}$ as the $C_{est}$. 
    
    Starting with $C_{est}$ as the first $C_{COP}$, the algorithm tries to find the next center candidate ($C_{cand}$). This is performed by $\textproc{ShortAdvancement}$, which attempts to move $C_{COP}$ by a fraction of a distance ($F_{pac}$) along the connecting vector between $C_{COP}$ and each contour point $h \in H_{part}$ using the following equation: $\overrightarrow{C_{cand}} = \overrightarrow{C_{COP}} + F_{pace} * (\overrightarrow{h} - \overrightarrow{C_{COP}})$. The first h resulting in a $C_{cand}$ where $R_{cand}>R_{COP}$, forces $C_{COP}$ to update to $C_{cand}$ and the remaining $h \in H_{part}$ are skipped to make way for the next iteration. A growing list of centers that have been visited ($C_{visited}$) is used to ensure no Euclidean distance (used to calculate $R_{cand}$) is unnecessarily performed because the resulting $C_{cand}$ will have already been visited and assessed from a past iteration. When $\forall h \in H_{part}$ the bubble does not move, Algorithm \ref{algorithm:BubbleGrowth} terminates.    
    
    \textbf{Bubble search:} Given $H_{all}$ and the $C_{COP}$, Algorithm \ref{algorithm:BubbleSearch} searches for WP ($W_1, W_2$). First, $R_{COP}$ is expanded to $R_{exp}$ (i.e., $R_{exp} = F_1 \times R_{COP}$) but continues expanding incrementally (i.e., $R_{exp} = F_2 \times R_{exp}$). $R_{exp}$ will expand, searching for a pair of contour points that exist inside of the expanded bubble (i.e., $h \in H_{inside}$ where $H_{inside} \subseteq H_{all}$). We say that $W_1$ and $W_2$ have been found if two contiguous $h \in H_{inside}$ meet the following criteria:
        \begin{itemize}
            \item[1.] $|\overline{W_1W_2}| > D_{min}$, where $D_{min} = F_3 \times R_{COP}$.
            \item[2.] $|\overline{W_1W_2}| < D_{max}$, where $D_{max} = F_4 \times R_{COP}$.
            \item[3.] $|\overline{M_{1,2}C_{ref}}| < |\overline{C_{COP}C_{ref}}|$, where $M_{1,2}$ is $\textproc{midpoint}(W_1,W_2)$.
        \end{itemize} 
        
    $D_{max}$ and $D_{min}$ are empirically determined limits of the distance between two WP in the scale of $R_{COP}$. 
    
    Algorithm \ref{algorithm:BubbleSearch} terminates if: (1) $W_1, W_2$ are found; or (2) $i_{max} = 10$ is reached; or (3) if before searching for $W_1$ and $W_2$ it is determined that $R_{COP} < |\overline{C_{COP}C_{ref}}|$. In case 2 and 3, $C_{Ref,edges}$ are selected as $W_1$ and $W_2$. Once WP are identified, all points defined by an inequality whose boundary passed through $W_1$ and $W_2$ and is opposite the side containing the COP is erased (pixels set to 0) to eliminate the forearm from the hand region. Finally, the border around the region is squeezed to the tightest bounding box, and the images is padded by 5 pixels and resized to $100 \times 100$. At this point, the hand region has been satisfactorily standardized as a CNN model input.
    
\vspace{-0.1in}
\subsection{Gesture Classification}
    A CNN model was trained to identify 10 pre-defined hand gestures. The architecture and the training parameters are summarized at the bottom of Figure \ref{figure:Overall Flowchart}. The model was trained using a \textit{categorical cross-entropy} loss function and an \textit{adam} optimizer with a variable learning rate.

    \begin{table} [b]
    \centering
        \caption{Average cost (sec) to detect hand region for different grid sizes and number of centroids; average computed over 7 contiguous frames.}
        \label{Table: Kmeans Clustering Times}
        \begin{tabular}{| c | c | c | c |c |}
            \hline
            \textbf{Centroids}      & \textbf{$5 \times 5$} & \textbf{$10 \times 10$} & \textbf{$15 \times 15$} & \textbf{$20 \times 20$} \\
            \hline
            2 only         & 0.0128 & 0.0161 & 0.0176 & 0.0235 \\
            2 to 3         & 0.0236 & 0.0298 & 0.0365 & 0.0397  \\
            2 to 4         & 0.0357 & 0.0451 & 0.0544 & 0.0576  \\
            2 to 5         & 0.0486 & 0.0624 & 0.0749 & 0.0822  \\
            \hline
        \end{tabular}
    \end{table}

\section{Experiments and Results}
    All experiments in this paper are performed on a desktop with 32GB of RAM, AMD Ryzen 7, 3700X, 8-core processor at 3.59 GHz on a 64-bit Windows 10 platform with build number 19043.1165. All CNN model training was performed through Google Co-laboratory using a GPU farm stationed at Google.
    
    For hand region detection, The number of centroids and the granularity of the grid in the silhouette analysis affects the cost of clustering. We observed the cost of finding an optimal number of centroids using silhouette analysis for different grid sizes on a single hand gesture over 7 frames. To keep cost low while also ensuring multiple hands are correctly isolated, the grid size should be fixed to a size of $10 \times 10$ and limited to at most 3 centroids. This limits the number of hands that can be detected in a single image to at most 3. 
    
    $|H_{part}|$ in \textit{bubble growth} is selected to strike a balance between the cost and accuracy. Table \ref{Table: BG Contour Point Analysis} justifies the selection of $|H_{part}|=30$. Experimentation has shown that $F_{pace}$ affects both the convergence speed and the accuracy of the center of palm. Larger values for $F_{pace}$ cause Algorithm \ref{algorithm:BubbleGrowth} to terminate pre-maturely, while lower value increases the computation cost. While $F_{pace} = 0.10$ worked best for us, an optimal value selected is left as a future work.

    \begin{table} [t]
        \centering
        \caption{Number of contour points in $H_{part}$ vs accuracy of Algorithm \ref{algorithm:BubbleGrowth} and cost (sec); number of bad bubbles reported as percentage of 291 images tested.}
        \label{Table: BG Contour Point Analysis}
        \begin{tabular}{| c | c | c |}
            \hline
            \textbf{Points} & \textbf{Bad Bubbles} & \textbf{Avg Cost}  \\
            \hline
            10        & 95.9\% & 0.0041  \\
            20        & 32.3\% & 0.0058  \\
            30        &  6.9\% & 0.0085  \\
            40        &  5.2\% & 0.0112  \\
            50        &  1.7\% & 0.0141  \\
            \hline
        \end{tabular}
        \vspace{-0.1in}
    \end{table}

    \begin{figure} [t]
        \includegraphics[width=\textwidth]{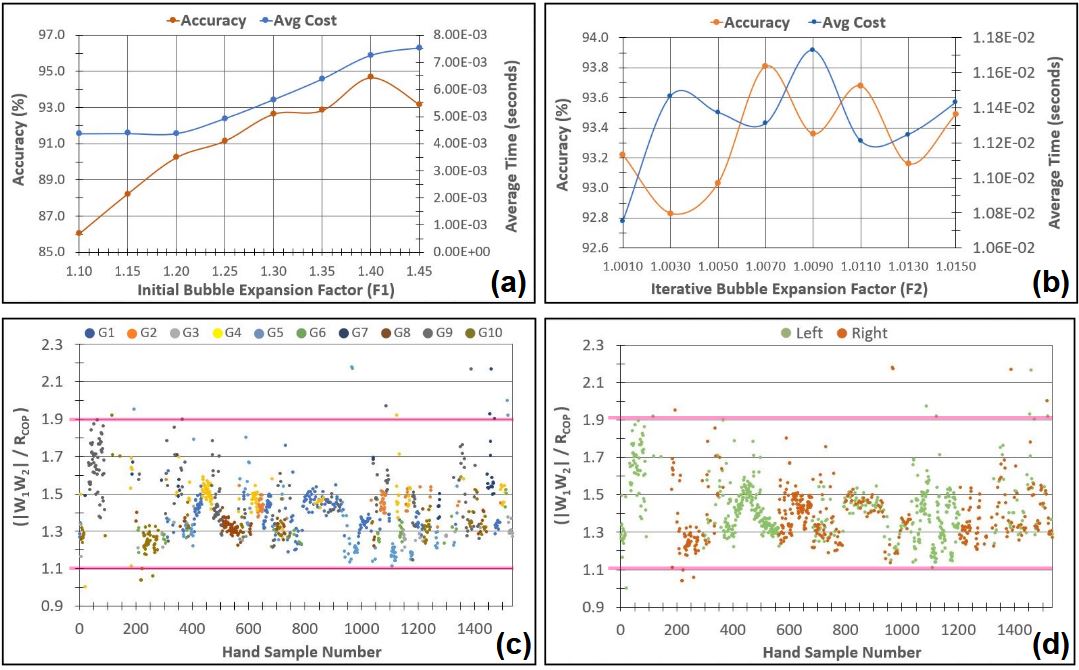}
        \caption{Plots to determine (a) $F_1$, (b) $F_2$, (c-d) $F_3$, and $F_4$.}
        \label{figure: F1 to F4}
        \vspace{-0.1in}
    \end{figure}

    A few experiments were performed to find optimize values for $F_1-F_4$ used in Algorithm \ref{algorithm:BubbleSearch} using 1000 images from 3 different users. A value of 1.2 has been proposed for factor $F_1$ in \cite{RealTimeHandGesture_2014}, but we propose $F_1=1.4$ after performing a study to assess the average accuracy and cost of Algorithm \ref{algorithm:BubbleSearch} keeping $F_2 = 1.01$ as constant. $F_1$ is varied between $(1.10, 1.45)$ in increments of 0.05 (Figure \ref{figure: F1 to F4}, (a)). Similarly, $F_2$ is varied between $(1.001, 1.015)$ in increments of 0.002 keeping $F_1 = 1.4$ as constant (Figure \ref{figure: F1 to F4}, (b)). Values of $F_3 = 1.1$ and $F_4 = 1.9$ are determined by plotting values of $\|\overline{W_1W_2}\|$ divided by $R_{COP}$ and selecting factors that bound 99\% of all plotted values(Figure \ref{figure: F1 to F4}, (c-d)). A value of $i_{max} = 10$ is determined such that it is small enough to minimize calculation time and large enough to allow searching for a significant number of hand samples.

    Bubble Growth and Bubble Search were tested using 1532 hand samples obtained using all 1,217 thermal images. The overall hand detection success rate of $95.64\%$ is on par with the $93.1\%$ to $98.7\%$ reported in \cite{ANNBased} and is detailed in Table \ref{Table: Overall Algorithm Metrics}. The term \textit{success} is defined for Algorithm \ref{algorithm:BubbleGrowth} as: (a) $C_{COP}$ exists in palm region; (b) $R_{COP} \approx R_{max-inscribed}$; (c) bubble contains $\leq 20$ black pixels. The term \textit{success} is defined for Algorithm \ref{algorithm:BubbleSearch} as: $W_1$ and $W_2$ are between $C_{COP}$ and $C_{Ref}$; (b) $\overline{W_1W_2}$ segments the hand and forearm regions as expected.

    \begin{table} [t]
        \centering
        \caption{Success and cost (sec/image) of Algorithm \ref{algorithm:BubbleGrowth} (BG) and Algorithm \ref{algorithm:BubbleSearch} (BS). Values listed overall (All) and by gesture number ($G_x$).}
        \label{Table: Overall Algorithm Metrics}
        \begin{tabular}{| c | c | c  c | c  c  c | c  c  c |}
            \hline
            \textbf{} & Samples & $BG_{success}$ & $BS_{success}$ & $BG_{avg}$ & $BG_{min}$ & $BG_{max}$ & $BS_{avg}$ & $BS_{min}$ & $BS_{max}$ \\
            \hline
            All       & 1532  &  95.64\% &  96.22\% &  0.012 &  0.001 & 0.120 &  0.007 &  2E-05 & 0.090 \\
            \hline
            $G_1$     & 325   &  96.92\% &  99.69\% &  0.009 &  0.002 & 0.030 &  0.004 &  0.003 & 0.060 \\
            $G_2$     & 64    &  98.48\% &  98.48\% &  0.014 &  0.001 & 0.030 &  0.006 &  0.001 & 0.040 \\
            $G_3$     & 67    &  98.51\% &  94.03\% &  0.010 &  0.003 & 0.030 &  0.008 &  0.005 & 0.070 \\
            $G_4$     & 153   &  100.0\% &  100.0\% &  0.012 &  0.005 & 0.040 &  0.004 &  2E-05 & 0.050 \\
            $G_5$     & 182   &  87.36\% &  92.86\% &  0.012 &  0.003 & 0.030 &  0.018 &  0.005 & 0.090 \\
            $G_6$     & 71    &  97.22\% &  98.61\% &  0.010 &  0.004 & 0.027 &  0.005 &  0.004 & 0.065 \\
            $G_7$     & 55    &  98.18\% &  94.55\% &  0.014 &  0.004 & 0.030 &  0.005 &  0.003 & 0.065 \\
            $G_8$     & 127   &  96.06\% &  99.21\% &  0.010 &  0.003 & 0.030 &  0.004 &  0.005 & 0.013 \\
            $G_9$     & 239   &  92.86\% &  99.58\% &  0.016 &  0.004 & 0.115 &  0.005 &  0.001 & 0.090 \\
            $G_{10}$    & 249   &  97.21\% &  86.85\% &  0.010 &  0.004 & 0.120 &  0.011 &  0.005 & 0.080 \\
            \hline
        \end{tabular}
    \end{table}

    The gesture classification model architecture was based on an LeNet-1 architecture. It was trained solely with the training data collected for this study and achieved a training accuracy of $99.9\%$. As seen in Table \ref{Table: Model Accuracies}, this model exhibits an overall testing accuracy of $96.9\%$ with our testing set. Our model was tested with \textit{finger-digit-05}; results are limited to 5 gestures because these were the only gestures matching those with which our model was trained. This is slightly better than $96.7\%$ reported in \cite{RealTimeHandGesture_2014}, which also used a model to classify 10 different gestures. This is also better than the $90.5\%$ recognition accuracy reported in \cite{Purple_Web_Hands}.
    
    We used data augmentation to randomly rotate hand samples between $0^{\circ}$ and $360^{\circ}$ to fit the model to hands at any orientation. Our process standardizes the size of hands when resizing the image, making our model zoom-agnostic. We experimented with batch sizes between 16 and 32 \cite{RealTime_MultiScale} to obtain the best training results and a variable learning rate between 1E-06 and 1E-04.  
    
    \begin{table} [t]
        \centering
        \caption{Accuracy of CNN gesture classification model overall ($G_{model}$) and by gesture number ($G_x$).}
        \label{Table: Model Accuracies}
        \begin{tabular}{| l | c | c | c | c | c | c | c | c | c | c | c |}
            \hline
            \textbf{}     & $G_{model}$ & $G_1$ & $G_2$ & $G_3$ & $G_4$ & $G_5$ & $G_6$ & $G_7$ & $G_8$ & $G_9$ & $G_{10}$ \\
            \hline
            Our Testing Set Accuracy       (\%) & 96.9 & 97.1 & 98.3 & 97.1 & 96.7 & 99.1 & 90.0 & 94.1 & 92.4 & 97.8 & 97.7  \\
            \hline
            finger-digit-05 Accuracy (\%) & 97.3 & 96.8 & 90.4 & - & 99.6 & 99.9 & - & - & - & 100 & -  \\
            \hline
        \end{tabular}
        \vspace{-0.1in}
    \end{table}

\vspace{-0.1in}
\section{Discussion and Conclusion}
    This paper proposes a real-time end-to-end system that can detect hand gestures from the video feed of a thermal camera. In that process the work introduces two novel methods for center of palm detection (bubble growth) and wrist point detection (bubble search) which are fast, accurate, and invariant to hand shape, hand orientation, arm length, and sizes (closeness to camera). 
    
    To maintain the real-time processing speed, our method can simultaneously detect hand gestures of 3 regions. However, this can be easily relaxed by introducing more processing power and distributed computing. This can also be achieved by using a heuristic to approximate the optimal number of centroids to use in clustering in lieu of testing a set of per-defined centroids using silhouette analysis. 
    
    Finally, we experimentally validated that our algorithm is user-agnostic (i.e. the algorithm can identify hand gestures of users that are not included in the training samples). Our system is highly accurate in detecting center of palm, wrist points, and hand gestures from hand masks produced from thermal images.
    
    Even though there are a lot of hand gesture detection algorithms available using color video data, only a few techniques solve the problem with thermal data. Our methods show that even though limited in features, thermal video is a viable medium to capture hand gestures for accurate gesture recognition. Moreover, future research can combine thermal with other data modalities (e.g., RGB, depth) for an even more robust hand gesture detection system.

\bibliography{OfficialBib}

\begin{thebibliography}{10}
\providecommand{\url}[1]{\texttt{#1}}
\providecommand{\urlprefix}{URL }
\providecommand{\doi}[1]{https://doi.org/#1}

\bibitem{ColorBased_TabletopInterfaces}
Bellarbi, A., Benbelkacem, S., Zenati, N., Belhocine, M.: Hand gesture
  interaction using color-based method for tabletop interfaces. In: 2011 IEEE
  7th International Symposium on Intelligent Signal Processing. pp.~1--6 (09
  2011). \doi{10.1109/WISP.2011.6051717}

\bibitem{RealTimeHandGesture_2014}
Chen, Z.h., Kim, J.t., Liang, J., Zhang, J., Yuan, Y.b.: Real-time hand gesture
  recognition using finger segmentation. The Scientific World Journal p. 9
  pages (May 2014). \doi{10.1155/2014/267872}

\bibitem{RealTime_BagOfFeatures}
Dardas, N.H., Georganas, N.D.: Real\-time hand gesture detection and
  recognition using bag\-of\-features and support vector machine techniques.
  IEEE Transactions on Instrumentation and Measurement  \textbf{60}, ~11
  (2011). \doi{10.1109/TIM.2011.2161140}

\bibitem{materialDetection}
Gately, J., Liang, Y., Wright, M.K., Banerjee, N.K., Banerjee, S., Dey, S.:
  Automatic material classification using thermal finger impression. In:
  MultiMedia Modeling. pp. 239--250. Springer International Publishing, Cham
  (2020)

\bibitem{HandLandmarks}
Grzejszczak, T., Kawulok, M., Galuszka, A.: Hand landmarks detection and
  localization in color images. Multimedia Tools and Applications  \textbf{75},
   16363–16387 (12 2015). \doi{10.1007/s11042-015-2934-5}

\bibitem{LayeredAchitecture}
Ibarguren, A., Maurtua, I., Sierra, B.: Layered architecture for real time sign
  recognition: Hand gesture and movement. Engineering Applications of
  Artificial Intelligence  \textbf{23}(7),  1216--1228 (2010).
  \doi{10.1016/j.engappai.2010.06.001}

\bibitem{ANNBased}
Islam, M.M., Siddiqua, S., Afnan, J.: Real time hand gesture recognition using
  different algorithms based on american sign language. In: 2017 IEEE
  International Conference on Imaging, Vision Pattern Recognition (icIVPR).
  pp.~1--6 (2017). \doi{10.1109/ICIVPR.2017.7890854}

\bibitem{In-Air}
Kim, S., Ban, Y., Lee, S.: Tracking and classification of in-air hand gesture
  based on thermal guided joint filter. Sensors  \textbf{17}(12), ~166 (Jan
  2017). \doi{10.3390/s17010166}

\bibitem{RealTime_MultiScale}
Meng, L., Li, R.: An attention-enhanced multi-scale and dual sign language
  recognition network based on a graph convolution network. Sensors
  \textbf{21}(4) (2021). \doi{10.3390/s21041120},
  \url{https://www.mdpi.com/1424-8220/21/4/1120}

\bibitem{FingerDigits}
O'Shea, R.: Finger digits 0-5 (11 2019),
  \url{https://www.kaggle.com/roshea6/finger-digits-05}

\bibitem{SilhouetteAnalysis}
Rousseeuw, P.J.: Silhouettes: A graphical aid to the interpretation and
  validation of cluster analysis. Journal of Computational and Applied
  Mathematics  \textbf{20},  53--65 (1987). \doi{10.1016/0377-0427(87)90125-7},
  \url{https://www.sciencedirect.com/science/article/pii/0377042787901257}

\bibitem{FastTrackingHandsFingertips}
Sato, Y., Kobayashi, Y., Koike, H.: Fast tracking of hands and fingertips in
  infrared images for augmented desk interface. In: Proceedings Fourth IEEE
  International Conference on Automatic Face and Gesture Recognition (Cat\.
  No\. PR00580). pp. 462--467 (2000). \doi{10.1109/AFGR.2000.840675}

\bibitem{AHD}
Song, E., Lee, H., Choi, J., Lee, S.: Ahd: Thermal image-based adaptive hand
  detection for enhanced tracking system. IEEE Access  \textbf{6},
  12156--12166 (2018). \doi{10.1109/ACCESS.2018.2810951}

\bibitem{FastRobust_DetectionOptimization}
Sridhar, S., Mueller, F., Oulasvirta, A., Theobalt, C.: Fast and robust hand
  tracking using detection-guided optimization. In: 2015 IEEE Conference on
  Computer Vision and Pattern Recognition (CVPR). pp. 3213--3221 (2015).
  \doi{10.1109/CVPR.2015.7298941}

\bibitem{Purple_Web_Hands}
Stergiopoulou, E., Papamarkos, N.: Hand gesture recognition using a neural
  network shape fitting technique. Engineering Applications of Artificial
  Intelligence  \textbf{22},  1141--1158 (2009).
  \doi{10.1016/j.engappai.2009.03.008}

\bibitem{DistanceTransform}
Strutz, T.: The distance transform and its computation (2021),
  \url{https://arxiv.org/abs/2106.03503}

\bibitem{PoseRecovery_UsingCNN}
Tompson, J., Stein, M., Lecun, Y., Perlin, K.: Real-time continuous pose
  recovery of human hands using convolutional networks. vol.~33 (08 2014).
  \doi{10.1145/2629500}

\bibitem{DeepDynamic_SegmentationRecognition}
Wu, D., Pigou, L., Kindermans, P.J., Le, N., Shao, L., Dambre, J., Odobez,
  J.M.: Deep dynamic neural networks for multimodal gesture segmentation and
  recognition. IEEE Transactions on Pattern Analysis and Machine Intelligence
  \textbf{38}, ~1--1 (03 2016). \doi{10.1109/TPAMI.2016.2537340}

\bibitem{COP_WristSegementation_2013}
Yao, Z., Pan, Z., Xu, S.: Wrist recognition and the center of the palm
  estimation based on depth camera. 2013 International Conference on Virtual
  Reality and Visualization pp. 100--105 (September 2013).
  \doi{10.1109/ICVRV.2013.24}

\bibitem{handTracking_LowCostHardware}
Yeo, H.S., Lee, B.G., Lim, H.: Hand tracking and gesture recognition system for
  human-computer interaction using low-cost hardware. Multimedia Tools and
  Applications  \textbf{74} (04 2013). \doi{10.1007/s11042-013-1501-1}

\bibitem{hand_pose_estimation}
Zhou, Y., Jiang, G., Lin, Y.: A novel finger and hand pose estimation technique
  for real-time hand gesture recognition. Pattern Recognition  \textbf{49},
  102--114 (2016). \doi{10.1016/j.patcog.2015.07.014}

\end{thebibliography}

\end{document}